\documentclass[10pt,twocolumn,letterpaper]{article}

\usepackage[pagenumbers]{cvpr}

\usepackage{graphicx}
\usepackage{amsmath}
\usepackage{amssymb}
\usepackage{booktabs}
\usepackage{url}
\usepackage{wrapfig}
\usepackage{times}
\usepackage{epsfig}
\usepackage{bm}
\usepackage{tabularx}
\usepackage{caption}
\usepackage{cite}
\usepackage{mathtools}
\usepackage{verbatim}
\usepackage{multirow}
\usepackage{array}
\usepackage{footnote}
\usepackage{color}
\usepackage{xcolor}
\usepackage{colortbl}
\usepackage[accsupp]{axessibility}
\newcommand{\Ours}[0]{$\textrm{D}^3\textrm{Former}$\xspace}

\newcommand{\customsubsubsection}[1]{%
  \par
  \pagebreak[2]%
  \refstepcounter{subsubsection}%
    \everypar={%
      {\setbox0=\lastbox}
      \addcontentsline{toc}{subsubsection}{%
        {\protect\makebox[0.3in][r]{\thesubsubsection.} \hspace*{3pt}#1}}%
      \textbf{\thesubsubsection\space\space{#1}\space}%
      \everypar={}%
    }%
  \ignorespaces
}

\usepackage[pagebackref,breaklinks,colorlinks]{hyperref}

\usepackage[capitalize]{cleveref}
\crefname{section}{Sec.}{Secs.}
\Crefname{section}{Section}{Sections}
\Crefname{table}{Table}{Tables}
\crefname{table}{Tab.}{Tabs.}

\begin{document}

\title{\Ours: Debiased Dual Distilled Transformer for Incremental Learning}

\author{%
Abdelrahman Mohamed$^{1 *}$ \quad Rushali Grandhe$^{1}$\thanks{Equal contribution} \\ \text{K J Joseph$^2$} \quad \text{Salman Khan$^{1,3}$} \quad \text{Fahad Khan$^{1,4}$} \\
$^1$Mohamed bin Zayed University of AI \quad $^2$Adobe Research \\ $^3$Australian National University \quad  $^4$ Linköping  University \\
}
\maketitle

\pagenumbering{arabic}   
\pagestyle{plain}

\begin{abstract}
   In class incremental learning (CIL) setting, groups of classes are introduced to a model in each learning phase. 
    The goal is to learn a unified model performant on all the classes observed so far. Given the recent popularity of Vision Transformers (ViTs) in conventional classification settings, an interesting question is to study their continual learning behaviour. In this work, we develop a Debiased Dual Distilled Transformer for CIL dubbed \Ours. The proposed model leverages a hybrid nested ViT design to ensure data efficiency and scalability to small as well as large datasets. In contrast to a recent ViT based CIL approach, our \Ours does not dynamically expand its architecture when new tasks are learned and remains suitable for a large number of incremental tasks. The improved CIL behaviour of \Ours owes to two fundamental changes to the ViT design. First, we treat the incremental learning as a long-tail classification problem where the majority samples from new classes vastly outnumber the limited exemplars available for old classes. To avoid the bias against the minority old classes, we propose to dynamically adjust logits to emphasize on retaining the representations relevant to old tasks. Second, we propose to preserve the configuration of spatial attention maps as the learning progresses across tasks. This helps in reducing catastrophic forgetting by constraining the model to retain the attention on the most discriminative regions. \Ours obtains favorable results on incremental versions of CIFAR-100, MNIST, SVHN,  and ImageNet datasets. Code is available at \url{https://tinyurl.com/d3former}. 
\end{abstract}

\section{Introduction}

Real world data is ever evolving and new object categories appear over time. Therefore, it is desired to learn models that can incrementally update their knowledge when the new data arrives, without forgetting the past concepts. Existing deep learning models \cite{lecun2015deep,schmidhuber2015deep} mostly consider a static world, where the learning happens once and if the model is trained on a new learning task, it catastrophically forgets the previously acquired knowledge \cite{kirkpatrick2017overcoming}. 

The goal of class incremental learning (CIL) is to continually learn new groups of classes (also referred to as tasks) without overwriting old task information \cite{joseph2022energy}. The main challenge is to balance the stability-plasticity trade-off, i.e., the model should be able to adapt to new tasks (plastic but not to the point of forgetting) while retaining past knowledge (stable but not leading to intransigence) \cite{abraham2005memory}.
The previous works mostly concentrate on convolutional neural networks (CNNs) in incremental learning settings  \cite{icarl,lucir,mnemonic,der}. However, self-attention \cite{transformer} based Vision Transformers (ViT) \cite{vit} have been shown to outperform CNNs on conventional classification settings \cite{khan2021transformers}. Therefore, understanding the capabilities of ViTs for CIL is an interesting and open research question.
\begin{figure*}
  \begin{minipage}[t]{.33\linewidth}
    \includegraphics[width=\linewidth]{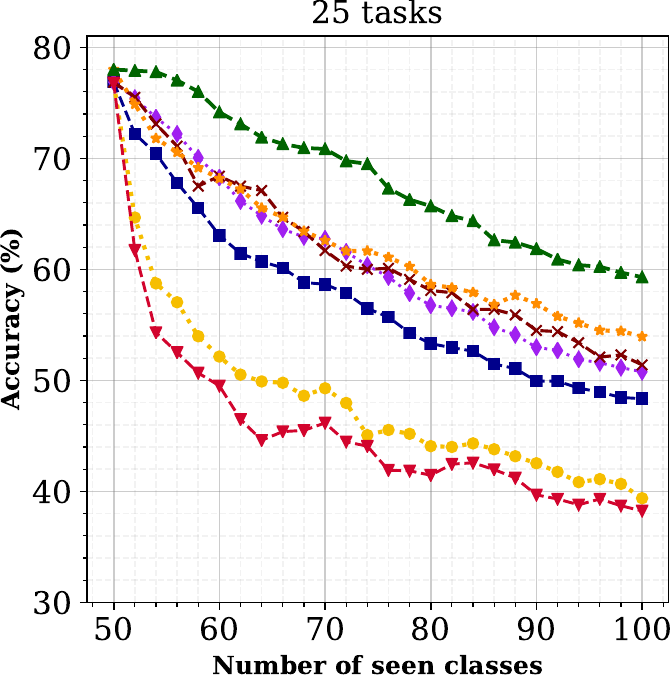}
  \end{minipage}\hfil
  \begin{minipage}[t]{.33\linewidth}
    \includegraphics[width=\linewidth]{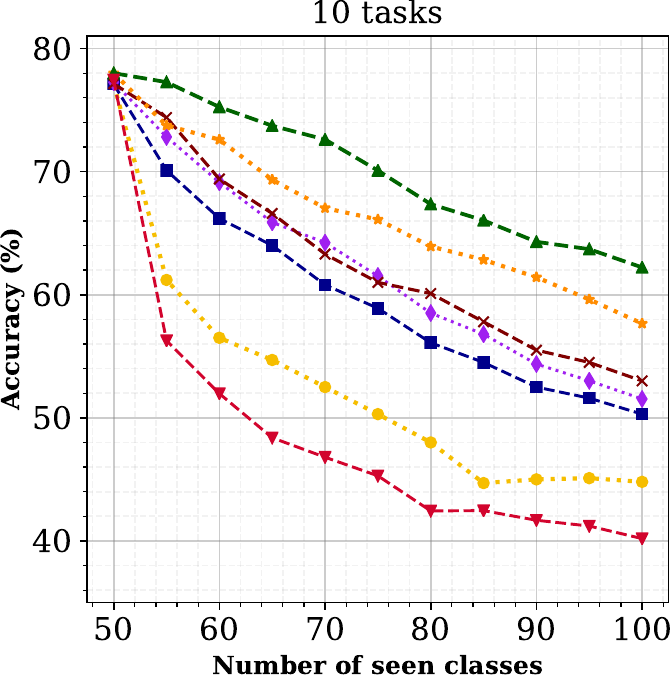}
  \end{minipage}\hfil
  \begin{minipage}[t]{.33\linewidth}
    \includegraphics[width=\linewidth]{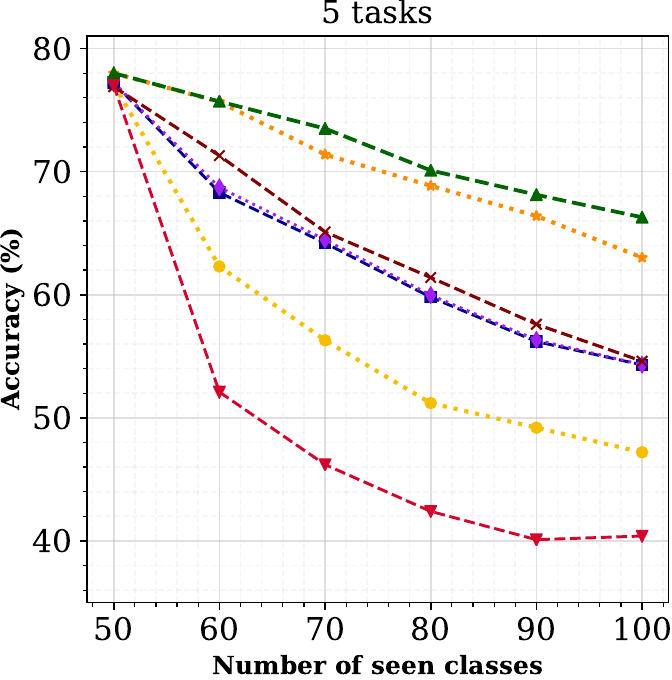}
  \end{minipage}
  \begin{minipage}[t]{0.98\linewidth}
    \includegraphics[width=\linewidth]{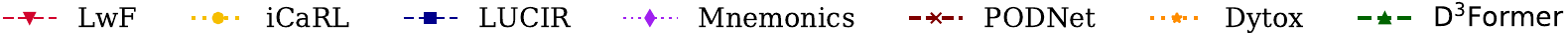}
  \end{minipage}
  \caption{\textbf{\Ours performance on small scale datasets:} Plots showing task wise accuracy for different number of incremental tasks for CIFAR-100.  \Ours achieves relatively high accuracy compared to other state-of-the-art methods when adding 2, 5 and 10 classes per task. 
 We present ImageNet-1K results in Tab. \ref{table_imagenet1k}, where we see a similar trend. Ours is the first transformer based incremental learning method, that scales well to small-scale and large-scale datasets alike.}
  \label{plot}
\end{figure*}

In this work, our goal is to develop a ViT model tailored for incremental learning settings. While ViTs have excelled in large data regimes, their plain versions lack the necessary inductive biases, thereby perform poorly on small datasets as compared to CNNs. This problem intensifies in incremental learning, where the new task dataset is generally much smaller than a typical full training set. A recent approach DyTox \cite{dytox} proposes the first incremental learning transformer model, however it has a dynamically expandable architecture which grows as the new tasks are learned.

We propose a hybrid ViT model for Incremental learning called \Ours (Debiased Dual Distilled Transformer). \Ours is data efficient and can be used equally well for both large and small-scale datasets (Fig.~\ref{plot}). The hybrid ViT designs \cite{swin,nest,vaswani2021scaling,cct} have proved to be more successful compared to pure self-attention based ViT designs at a lower computational cost. Specifically, our approach is based on a Nested Vision Transformer \cite{nest}, that uses local self-attention within the patches and then hierarchically aggregates non-local information via convolution and pooling operations. The benefit manifests via improved data efficiency which is important for the incremental training where each task episode has a limited data belonging to a relatively small group of classes.

In order to render the ViT amenable to incremental setting, we propose two key components to minimize catastrophic forgetting.
(a)  \emph{Debiasing via Logit Adjustment:} In the incremental phases, usually a small exemplar set of old task data is maintained due to memory constraints \cite{icarl}. Since the classes in exemplar set are heavily imbalanced w.r.t the new task data, it bias the model against the previously observed classes. We propose a simple logit adjustment strategy to put appropriate emphasis on the previous task classes to avoid representational and classifier biasness.
(b) \emph{Dual Distillation:} In addition to the regular distillation loss applied on the logits / features \cite{icarl,lucir}, we propose to maintain the attention cast on the input image by the teacher model and the student model to be consistent as the incremental learning progresses. To this end, we leverage the visual interpretability properties of Nested Transformer \cite{nest} to obtain salient regions using simple methods such as gradient based class activation maps, that are enforced to be consistent during incremental learning. 

\noindent In summary, the main highlights of our approach are:
\begin{itemize}\setlength{\itemsep}{0em}
    \item We develop the first hybrid Transformer model for incremental settings, that can adaptively learn new task distributions.  In comparison to state of the art methods \cite{der,dytox,rps}, our approach performs favorably well, as shown in Fig. \ref{plot}, even without dynamically expanding its parameters as the number of tasks grow, making it scale easily.
    \item Owing to the inherent  long tail distribution in CIL, our debiased loss formulation allocates high emphasis to the imbalanced data from old tasks, thereby minimizing loss of information relevant to previous tasks.
    \item We show that maintaining the attention on regions that are most crucial for predicting a particular class helps avoid overwriting the important features during incremental learning.  
    \item Our extensive results on CIFAR-100, MNIST, SVHN and ImageNet datasets demonstrate considerable gains over the recent top performing incremental learning methods in terms of average and final task accuracies, as well as minimizing the forgetness measure. 
\end{itemize}

\section{Related Work}
\subsection{Incremental Learning}

We focus on class-incremental learning, where new classes are introduced to the model in distinct training phases. The methods are usually grouped into the following heads -

\noindent \emph{\bf{Regularization based:}}
Knowledge distillation \cite{hinton2015distilling} has been extensively used as a regularizer to minimise the changes to the decision boundaries of previous classes while learning incrementally.
The model trained until the earlier phases of learning is treated as a teacher network, whose penultimate features or the logits are distilled into the incremental model.
This was introduced in LwF~\cite{lwf} and has been widely adopted by later methods.
iCaRL \cite{icarl} uses KL Divergence loss for knowledge distillation. LUCIR \cite{lucir} uses cosine similarity based loss for knowledge distillation and margin ranking loss for the hard examples. PODNet \cite{pod} uses pooling as a means of restricting change. LwM \cite{lwm} and RRR \cite{rrr} encourages the model to remember by making use of explanability techniques.

\noindent \emph{\bf{Replay based: }}In memory replay based methods, a small subset of data from the older classes are retained and replayed while learning the later incremental phases. This helps to alleviate the distributional shift caused while learning the new classes~\cite{icarl,aanet,lucir}. 
Examples to be stored in the replay buffer may be randomly selected across all tasks \cite{riemer2018learning,wu2019large}, randomly selected per task \cite{joseph2021open,kj2021incremental}, by selecting an optimal coreset based on gradient statistics \cite{tiwari2021gcr} or even by solving a submodular objective \cite{brahma2018subset}. 
An alternative to storing exemplars would be to learn the distribution of the data using generative models and replaying the generated pseudo images \cite{shin2017continual}. We refer reader to \cite{rehersal} for a more detailed treatment on replay-based continual learning methods. Replay based methods undesirably introduce bias towards new classes  due to class imbalance. BiC \cite{bic} learns an MLP explicitly to correct the bias, while SS-IL \cite{ssil} uses task-wise distillation along with separate heads for the current and previous tasks. \cite{jodelet2021balanced} proposes balancing softmax outputs to reduce bias.

\noindent \emph{\bf{Structure based:}}
Structure based methods usually allocate additional parameters for every new incremental phase. RPSNet \cite{rps}, learns different paths for different tasks, ensuring weight sharing among tasks. DER \cite{der} adds a new feature extractor for every task and uses pruning to reduce model size. A recent ViT based method DyTox \cite{dytox}, proposes to use a dynamic task-token expansion based method to facilitate incremental learning.

\subsection{Vision Transformers}
Self-attention based Transformer architecture \cite{transformer} has revolutionized NLP. Vision Transformer (ViT) \cite{vit} has helped to carry-over the successes from the NLP community to computer vision. Some of the notable ViT architecture include DeiT \cite{deit} which uses knowledge distillation from a convolutional neural network through a distillation token, T2T ViT \cite{t2t} which tries to preserve local structure and reduce number of tokens by aggregating neighbouring tokens, XCiT \cite{xcit} which performs self-attention across feature channels to counter the quadratic complexity associated with self-attention between tokens.
Recently, several hybrid ViTs -- which use convolution layers along with self-attention -- have been introduced. 
CvT \cite{cvt} , CCT \cite{cct}, Swin \cite{swin} and Nested Transformer (NesT) \cite{nest} are among the popular hybrid ViTs. To the best of our knowledge, ours is the first method that makes use of a hybrid ViT architecture for continual learning.

\section{\Ours: Debiased Dual Distilled Transformer}

\begin{figure*}[htp]
    \centering
    \includegraphics[width=1\textwidth]{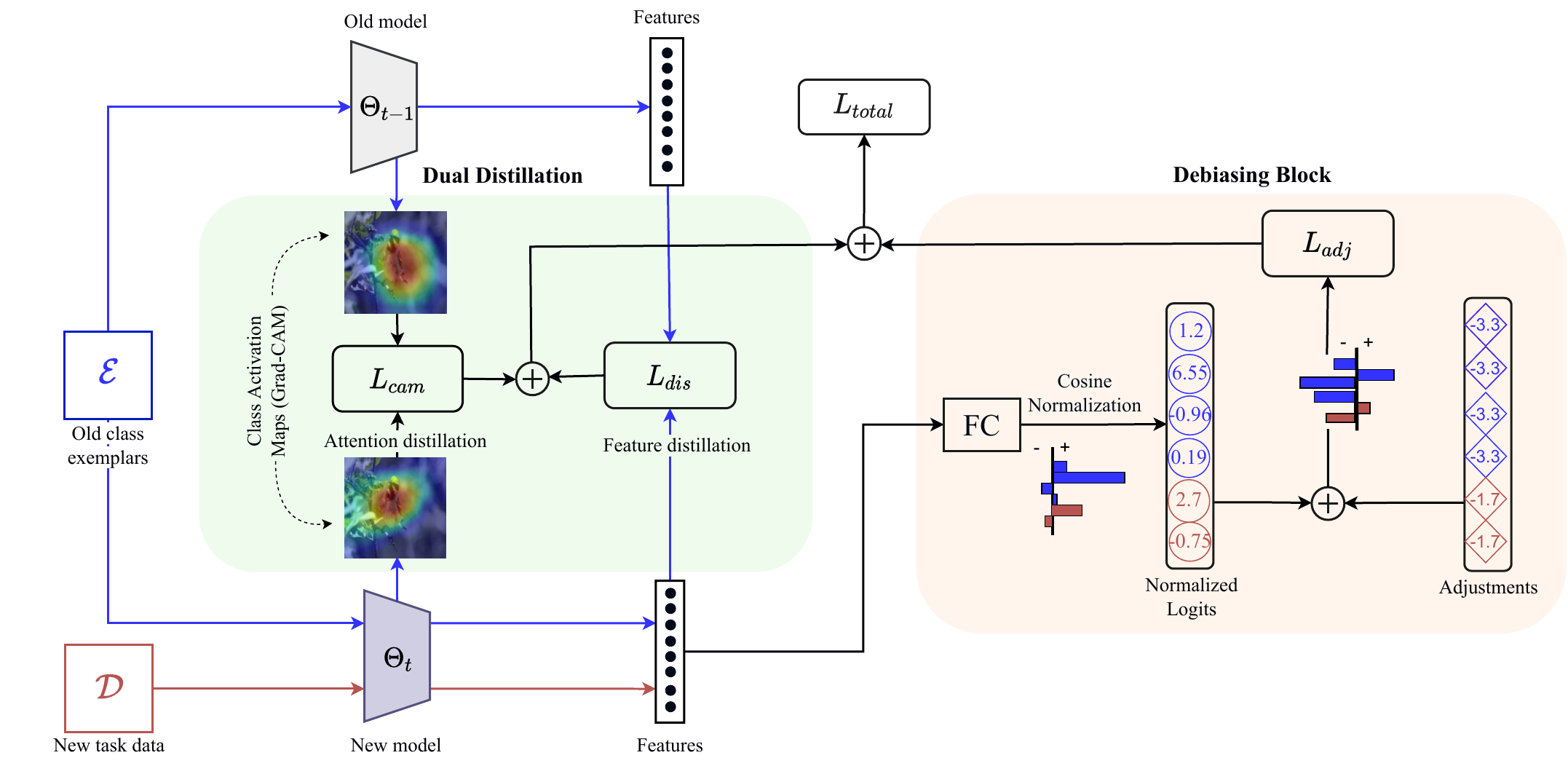}
    \caption{\textbf{(a)} Dual distillation: \emph{(left)} In each learning phase $t$, the previous phase model $\bm{\Theta_{t-1}}$ is used to extract the features and Grad-CAMs of exemplars $\mathcal{E}$. Later, these attention maps are compared with the current model $\bm{\Theta_t}$ attention maps and $L_{cam}$ loss is calculated between them. It discourages changes to the spatial attention response of $\bm{\Theta_t}$ w.r.t old classes. Knowledge distillation loss ($L_{dis}$) is computed as the cosine similarity between the features of $\mathcal{E}$ from $\bm{\Theta_{t-1}}$ and $\bm{\Theta_t}$. This maintains the orientation of the feature vectors for the old classes. \textbf{(b)} Debiasing block: (\emph{right})  To compensate for bias towards new classes, in addition to cosine normalization of the logits, adjustments are added to the logits before applying cross-entropy. The adjusted logits result in stronger updates for the old (rare) classes to avoid their misclassification, thereby minimizing forgetting old task knowledge.}
    \label{fig:arch}
\end{figure*}

Incrementally learning a classifier to expand its knowledge, without hampering its performance on the earlier set of classes is an arduous task for deep learning models. In our work, \Ours, we propose to make use of a hybrid model -- that utilises the complementary advantages of transformer architectures and convolutional network -- for class incremental learning. We detail about the model architecture in Sec.~\ref{3.1}.
Exemplar replay has emerged as a simple yet effective method to alleviate forgetting. Due to storage limitations, we store only few examples (close to 20 examples per task) in the exemplar memory. While learning a new task, we combine the data from exemplar memory with the incoming data. This skews the training data towards the latest task. We propose to address this imbalance by treating this setting as a long-tailed recognition problem, as explained in Sec.~\ref{3.2}.
Further, in Sec.~\ref{3.3}, we propose to retain the spatial attention of exemplar images across tasks. This has a two fold effect: firstly, it improves the spatial awareness of the model; secondly, it helps to reduce forgetting by reminding the model on how it needs to attend to the more discriminative parts of the images during incremental learning.
Concretely, let us consider learning a model $\bm{\Theta}$ across a total of $N+1$ training phases, where the first phase ($t=0$) involves learning a set of $B$ base classes, followed by $N$ incremental phases. Each phase ($1 \leq t \leq N$) involves learning a fixed number of $C$ new classes.
Consider the number of exemplars retained for each class in the previous tasks ($0 \ldots t-1$) is $M$, thereby forming a set $\mathcal{E} = \{\mathcal{E}_0, ..\mathcal{E}_{t-1}\}$.  Thus, in the incremental phases, the model $\bm{\Theta}$ is trained using the replayed old class exemplars $\mathcal{E}$ and all new input classes data $\mathcal{D}$.
Figure \ref{fig:arch} illustrates the overall setup and the different loss functions used in \Ours. $L_{cam}$ and $L_{dis}$ enforces the current model to not deviate much from the previous model, while a cross entropy loss on the adjusted logits ($L_{adj}$) helps to learn the new task. We explain more on these in the following sub-sections.

\subsection{The Hybrid ViT Model}\label{3.1}
\begin{figure*}[htp]
    \centering
    \includegraphics[width=0.95\textwidth]{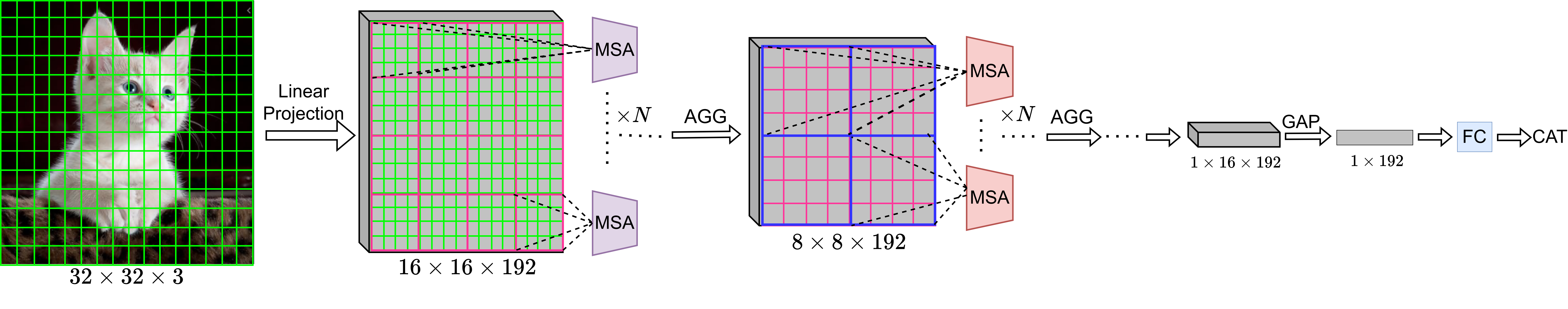}
    \caption{Nested Transformer (NesT) architecture illustrating blockify and aggregation operations. }
    \label{fig:nest}
\end{figure*}

\Ours builds upon the hybrid ViT NesT \cite{nest} which makes use of 2 basic operations - blockify and aggregation. The blockify operation combines spatially adjacent embeddings into a group. It captures intra-block information or local attention using several stacked transformer encoders. Each transformer encoder consists  of Layer Normalization (LN) and Multi-head self-attention (MSA) followed by Feed-Forward network (FFN). On the other hand, the aggregation operation (AGG) combines neighboring blocks with the help of a simple convolution and pooling layer. It captures inter-block relationships and helps gain global understanding of an image. The local and global processing steps help learn discriminative  features.\\
The above operations are repeated alternately to eventually create the hierarchical structure of NesT (Fig.~\ref{fig:nest}), where each hierarchy shares the same set of parameters. The final class prediction is performed through a global average pooling (GAP) layer followed by a fully connected (FC) layer. NesT is  characterized by two parameters, patch size, $S$ and number of block hierarchies, $T_d$. To render NesT suitable for CIL, we propose two principal modifications - Debiasing via Logit Adjustment and Dual Distillation.

\subsection{Reducing the Bias in the Logits}\label{3.2}

In incremental phases, a small set of exemplars are usually stored for old tasks data due to memory constraints. However, current task samples outnumber old tasks exemplars in each phase leading to a strong bias towards new classes.\\
An intuitive approach to reduce bias involves placing more emphasis on rare classes during the learning process. This can be easily implemented using a simple logit adjustment strategy \cite{logit}. Logit adjustment adds an appropriate offset to the output logits thereby increasing the margin between rare and frequent classes. The offset can be calculated as $\tau \log \pi_y$, where $\tau$ is a hyperparameter that controls the adjustment strength, $\pi_y$ is the estimated prior for class $y$.
Class priors are approximated as the frequency of each class in the dataset. However, in our case, since the number of exemplars from the old classes are equal and the number of samples from each new classes samples are also equal, there needs to be only two class priors $\{\pi_o, \pi_n\}$. The class priors for old and new classes are calculated as follows: 
\begin{align}
    \pi_o = \frac{|\mathcal{E}_{c_o}|}{|\mathcal{E}| + |\mathcal{D}|}, \forall c_o \in \mathcal{E}  , \,\, \pi_n = \frac{|\mathcal{D}_{c_n}|}{|\mathcal{E}| + |\mathcal{D}|}, \forall c_n \in \mathcal{D}
\end{align}
Where $c_o$ are the old classes, and $c_n$ are the new classes. 
Thus, the cross-entropy loss can be modified by including the logit adjustment offsets as:
\noindent\begin{equation}
\begin{split}
    L_{adj}(\bm{x}) = - \log \frac{e^{f_y(\bm{x})+\tau \log \pi_y}}{\displaystyle\sum_{y' \in \mathcal{T}} e^{f_{y'}(\bm{x})+\tau \log \pi_{y'}}}, \;\; \\ s.t., \ \pi_y, \pi_{y'} \in \{\pi_o, \pi_n\},
    \label{eq1}
\end{split}
\end{equation}
where $\mathcal{T}$ is the class labels set, and $f_y(\bm{x})$ is the cosine normalized logits for an input sample $\bm{x}$. Cosine normalization helps in further reducing bias towards new classes samples \cite{lucir,cos}, and computed as:
\noindent\begin{equation}
    f_y(\bm{x}) = \eta\langle\bar{\theta}(\bm{x}),\bar{w}\rangle,
    \label{eqcosine}
\end{equation}
where $\eta$ is a learnable scaling parameter to control the peakness of the logits for softmax, as the values after normalization are between $[-1,1]$, $\bar{\theta}(\bm{x})$  is the $L_2$-normalized extracted features, and $\bar{w}$ denotes the final layer $L_2$-normalized weights.

\begin{table*}[t]

  \centering
  \vspace{-0.2cm}
    \resizebox{\textwidth}{!}{
  \begin{tabular}{p{3.5cm}p{1.5cm}p{1.5cm}p{1.0cm}p{0.5cm}p{1.5cm}p{1.5cm}p{1.0cm}p{0.5cm}p{1.5cm}p{1.5cm}p{1.0cm}}
  \toprule
   \multirow{3}{*}{Method} & \multicolumn{3}{c}{\emph{$N$=5}} && \multicolumn{3}{c}{\emph{$N$=10}} && \multicolumn{3}{c}{\emph{$N$=25}}\\
  \cmidrule{2-4} \cmidrule{6-8} \cmidrule{10-12}
   & Avg $\uparrow$ & Last $\uparrow$ & $\mathcal{F}$ $\downarrow$ && Avg $\uparrow$ & Last $\uparrow$ & $\mathcal{F}$ $\downarrow$ &&  Avg $\uparrow$ & Last $\uparrow$ & $\mathcal{F}$ $\downarrow$\\
    \midrule
    \midrule
    LwF\cite{lwf} & 49.59 & 40.40 & 43.36 && 46.98 & 40.19 & 43.58 && 45.51 & 38.25 & 41.66\\
    BiC\cite{bic} & 59.36 & - & 31.42 && 54.20 & - & 32.50 && 50.00  & -  & 34.60\\

    iCaRL \cite{icarl} & 57.12\tiny{$\pm$ 0.50} & 47.20 & 31.88 && 52.66\tiny{$\pm$ 0.89} & 44.80 & 34.10 && 48.22\tiny{$\pm$ 0.76} & 39.39 & 36.48\\

    LUCIR \cite{lucir} & 63.17\tiny{$\pm$ 0.87} & 54.30  & 18.70  && 60.14\tiny{$\pm$ 0.73} & 50.30 & 21.34 && 57.54\tiny{$\pm$ 0.43} & 48.35 & 26.46\\

    Mnemonics \cite{mnemonic} & 63.34\tiny{$\pm$ 0.34} & 54.32 & \textbf{10.91} && 62.28\tiny{$\pm$ 0.61} & 51.53 & \textbf{13.38} && 60.96\tiny{$\pm$ 0.72} & 50.78 & \textbf{19.80}\\
    PODNet-CNN \cite{pod} & 64.83\tiny{$\pm$ 1.11} & 54.60 & -  && 63.19\tiny{$\pm$ 1.31}  & 53.00  &  - && 60.72\tiny{$\pm$ 1.54}  & 51.40 & -\\
 \midrule
    DyTox$ ^\ast$ \cite{dytox} & 70.28 & 63.02 & 24.54  && 66.72  & 59.62  &  29.86 && 62.83 & 53.95 & 33.72\\
    \rowcolor{blue!6} \textbf{\Ours (ours)} & \textbf{72.23}\tiny$\pm0.08$ & \textbf{66.24}\tiny$\pm0.1$ & \underline{12.09}  && \textbf{70.94}\tiny$\pm 0.43$  & \textbf{63.10}\tiny$\pm0.54$  & \underline{16.12}  &&  \textbf{68.68}\tiny$\pm0.4$  & \textbf{59.79}\tiny$\pm0.44$  & \underline{21.23}\\
    \rowcolor{blue!6} \textbf{\Ours-NCM (ours)} & \underline{71.38}\tiny$\pm 0.32$ & \underline{64.26}\tiny$\pm0.47$ & 16.52  && \underline{69.35}\tiny$\pm0.47$  & \underline{61.46}\tiny$\pm0.58$ & 19.36  &&  \underline{67.03}\tiny$\pm0.59$  & \underline{58.12}\tiny$\pm0.80$  & 22.84\\
  \bottomrule
\end{tabular}}
   \caption{
  Results of \textbf{CIFAR-100} with Average accuracy (\%), last phase accuracy (\%) and forgetting rate $\mathcal{F}$(\%) of different methods in 5,10 and 25 tasks settings. The top group of methods are based on CNN while the last three approaches (including ours) are based on transformer models. $ ^\ast$ indicates results reproduced by us using author's official codebase.
}
 \label{table_cifar}
\end{table*}

\subsection{Dual-distillation Framework}
\label{3.3}

Knowledge distillation was introduced to CIL  as a means of reducing forgetting by transferring knowledge about old tasks from the teacher model $\bm{\Theta_{t-1}}$ to the student model $\bm{\Theta_t}$ \cite{lwf,icarl}. 
First, we incorporate knowledge distillation at feature-level \cite{lucir} using a cosine similarity loss based on feature vectors  computed as follows :
\begin{equation}
    L_{dis} = 1 - \langle \bar{\theta}_{t-1}(\bm{x}), \bar{\theta}_t(\bm{x}) \rangle,
    \label{eq:dis}
\end{equation}
where $\bar{\theta}_{t-1}(\bm{x})$, $\bar{\theta}_t(\bm{x})$ denote the normalized feature vectors extracted from models $\bm{\Theta_{t-1}}$ and $\bm{\Theta_t}$, respectively.\\
In addition to Eq. \ref{eq:dis} which preserves the orientation of feature vectors as incremental learning progresses, preserving the model response on regions that are critical for predicting a particular class can help further reduce catastrophic forgetting \cite{lwm,rrr}. In contrast to pure self-attention based ViTs, the enhanced visual interpretability of hybrid ViTs \cite{nest, swin} allows us to extract these salient regions utilizing general methods such as gradient based class activation maps (Grad-CAM) \cite{chefer2021transformer, cam}. Grad-CAMs are essentially the heatmaps which localize the most discriminative regions for a particular class in a given image. We enforce that the attention response of $\bm{\Theta_{t}}$ on the old tasks must be maintained similar to that of $\bm{\Theta_{t-1}}$ through a Grad-CAM based $L_1$ distillation loss:
\begin{equation}
    L_{cam}(x) = \parallel CAM(\bm{\Theta_{t}}, \bm{x}) - CAM(\bm{\Theta_{t-1}},\bm{x}) \parallel_1
    \label{eq:cam}
\end{equation}

We obtain Grad-CAMs from the feature maps of the final hierarchy in NesT, since it contains global information of the whole image. The total loss can thus be written as:
\begin{equation}
\begin{split}
    L_{total} = \frac{1}{|\mathcal{T}|} \displaystyle\sum_{y \in \mathcal{T}} L_{adj}(\bm{x}) + \frac{\lambda}{|\mathcal{N}_o|} \displaystyle\sum_{y \in \mathcal{N}_o} L_{dis}(\bm{x}) \\ + \frac{\gamma}{|\mathcal{N}_o|} \displaystyle\sum_{y \in \mathcal{N}_o} L_{cam}(\bm{x}),
    \label{eq:total}
\end{split}
\end{equation}
where $y$ is the class label of sample $\bm{x}$, $\mathcal{N}_o$ denotes the set of old classes, $\mathcal{T}$ is the set of all classes, $\lambda$ is a scaling factor controlling cosine similarity based knowledge distillation and $\gamma$ is a scaling factor controlling the magnitude of Grad-CAM based distillation. Note that unlike \cite{lwm}, Eq. \ref{eq:total} employs Grad-CAM distillation only on exemplars rather than all training samples. This led to improved training stability as it directs the model's attention to regions relevant to old classes rather than potentially distracting regions encountered when learning new classes. Related results are also shown in Table \ref{table_mix}.

\setlength{\tabcolsep}{1.0mm}{

\begin{table*}[ht]

  \centering
  \vspace{-0.2cm}
  \resizebox{\textwidth}{!}{
  \begin{tabular}{p{3.5cm}p{1.5cm}p{1.5cm}p{1.0cm}p{0.5cm}p{1.5cm}p{1.5cm}p{1.0cm}p{0.5cm}p{1.5cm}p{1.5cm}p{1.0cm}}
  \toprule
   \multirow{3}{*}{Method} & \multicolumn{3}{c}{\emph{$N$=5}} && \multicolumn{3}{c}{\emph{$N$=10}} && \multicolumn{3}{c}{\emph{$N$=25}}\\
  \cmidrule{2-4} \cmidrule{6-8} \cmidrule{10-12}
   & Avg $\uparrow$ & Last $\uparrow$ & $\mathcal{F}$ $\downarrow$ && Avg $\uparrow$ & Last $\uparrow$ & $\mathcal{F}$ $\downarrow$ &&  Avg $\uparrow$ & Last $\uparrow$ & $\mathcal{F}$ $\downarrow$\\
    \midrule
    \midrule
    DyTox Joint & - & 79.82 & - && - & 79.82 & - && - &  79.82 & -\\
    \Ours Joint & - & 82.14 & - && - & 82.14 & - && - & 82.14  & -\\
    LwF \cite{lwf} & 53.62 & 40.10 & 55.32 && 47.64 & 36.10 & 57.00 && 44.32 & 34.12 & 55.12\\
    BiC \cite{bic} & 70.07 & - & 27.04 && 64.96 & - & 31.04 && 57.73  & -  & 37.88\\

    iCaRL \cite{icarl} & 65.44\tiny{$\pm$ 0.35} & 53.60 & 43.40 && 59.88\tiny{$\pm$ 0.83} & 49.10 & 45.84 &&  52.97\tiny{$\pm$ 1.02} & 43.34 & 47.60\\

    LUCIR \cite{lucir} & 70.84\tiny{$\pm$ 0.69} & 60.00  &  31.88 && 68.32\tiny{$\pm$ 0.81} & 57.10 & 33.48 && 61.44\tiny{$\pm$ 0.91} & 49.26 & 35.40\\

    Mnemonics \cite{mnemonic} & 75.54\tiny{$\pm$ 0.85} & 61.36 & \textbf{17.40} && 74.33\tiny{$\pm$ 0.56} & 59.56 & \textbf{17.08} && 68.31\tiny{$\pm$ 0.39} & 59.22 & \textbf{20.83}\\
    PODNet-CNN \cite{pod} & 76.96\tiny{$\pm$ 0.29} & 67.60 &  - && 73.70\tiny{$\pm$ 1.05}  & 65.00  & -  && 71.78\tiny{$\pm$ 2.77}  & 54.30 & -\\
    \midrule
    DyTox$ ^\ast$ \cite{dytox} & 77.08 & \textbf{70.24} & 21.21  &&  74.06 & \textbf{65.44}  & 27.16   && 68.76 & \textbf{61.54} & 30.04 \\
    \rowcolor{blue!6}
    \textbf{\Ours (ours)} & \textbf{77.31}\tiny{$\pm$ 0.41} & 67.82\tiny{$\pm$ 0.36} & 25.92 &&  \underline{75.01}\tiny{$\pm$ 0.63} & 63.46\tiny{$\pm$ 0.32}  & 27.41  && \textbf{72.43}\tiny{$\pm$ 0.76}  & 59.91\tiny{$\pm$ 1.1} & 30.80 \\
    \rowcolor{blue!6}
    \textbf{\Ours-NCM (ours)} & \underline{77.21}\tiny{$\pm$ 0.22} & \underline{69.89}\tiny{$\pm$ 0.18} & \underline{17.98} &&  \textbf{75.26}\tiny{$\pm$ 0.28} & \underline{65.11}\tiny{$\pm$ 0.25}  & \underline{20.21}  && \underline{72.31}\tiny{$\pm$ 0.24}  & \underline{60.01}\tiny{$\pm$ 0.85} & \underline{27.20}\\
  \bottomrule
\end{tabular}}
   \caption{
  Results of \textbf{Imagenet-100} with Average accuracy (\%), last phase accuracy (\%) and forgetting rate $\mathcal{F}$ (\%) of different methods in 5,10 and 25 task settings. $ ^\ast$ indicates results reproduced by us using author's official codebase.
}
 \label{table_imagenet100}
\end{table*}
}

\begin{table*}[ht]
\centering
\resizebox{0.95\textwidth}{!}{
\begin{tabular}{p{5.2cm}p{2cm}p{1.5cm}p{1.5cm}p{0.5cm}p{2cm}p{1.5cm}p{1.5cm}}
\toprule
\multirow{3}{*}{Method} & \multicolumn{3}{c}{\emph{$N$=5}} && \multicolumn{3}{c}{\emph{$N$=10}}\\
\cmidrule{2-4} \cmidrule{6-8}
& Avg $\uparrow$  & Last $\uparrow$ & $\mathcal{F}$ $\downarrow$ && Avg $\uparrow$ & Last $\uparrow$ & $\mathcal{F}$ $\downarrow$\\
\midrule
\midrule
DyTox Joint & - & 73.58 & - && - & 73.58 & - \\
\Ours Joint & - & 76.42 & - && - & 76.42 & - \\
LwF~\cite{lwf} & 44.35 & 34.20 & 48.70 && 38.90 & 30.10 & 47.94\\
BiC~\cite{bic} & 62.65 & - & 25.06 && 58.72 & - & 28.34\\

iCaRL \cite{icarl} & 51.50 \tiny{$\pm$ 0.43} & 34.20 & 26.03 && 46.89 \tiny{$\pm$ 0.35} & 38.91 & 33.76\\

LUCIR \cite{lucir} & 64.45 \tiny{$\pm$ 0.32} & 56.60  &  24.08 && 61.57 \tiny{$\pm$ 0.23} & 51.7 & \underline{27.29}\\

Mnemonics \cite{mnemonic} & 64.54 \tiny{$\pm$ 0.49} & 56.85 & \textbf{13.85} && 63.01\tiny{$\pm$ 0.57} & 54.99 & \textbf{15.82}\\
PODNet-CNN \cite{pod} & 66.43 & 58.90 &  - && 63.21 & 55.70 & -\\
\midrule
\rowcolor{blue!6}
DyTox$ ^\ast$ \cite{dytox} & 68.96 & 64.08 &  18.63 && 67.12 & 57.61 & 31.83\\
\rowcolor{blue!6}
\textbf{\Ours (ours)} & \textbf{72.73}\tiny{$\pm$ 0.30} & \underline{64.58}\tiny{$\pm$ 0.33} & 21.41 && \underline{69.56}\tiny{$\pm$ 0.29} & \underline{59.22}\tiny{$\pm$ 0.32} & 32.35\\
\rowcolor{blue!6}
\textbf{\Ours-NCM (ours)} & \underline{72.61} \tiny{$\pm$ 0.32} & \textbf{64.64} \tiny{$\pm$ 0.29} &  \underline{17.03} && \textbf{70.04}\tiny{$\pm$ 0.34} & \textbf{59.90}\tiny{$\pm$ 0.31} & 27.87\\
  \bottomrule
\end{tabular}}
\caption{
  Results of \textbf{ImageNet-1K} with Average accuracy (\%), last phase accuracy (\%) and forgetting rate $\mathcal{F}$ (\%) of different methods in 5 and 10 tasks setting. $ ^\ast$ indicates results reproduced by us using author's official codebase.
}
\label{table_imagenet1k}
\end{table*}

\section{Experiments}
\label{sec:exp}
We analyze the performance of \Ours on large scale datasets such as ImageNet-1K\cite{ILSVRC15}, ImageNet Subset-100  and small scale datasets like MNIST\cite{mnist}, SVHN\cite{svhn} and CIFAR-100\cite{CIFAR-100}.
MNIST contains $28 \times 28$ pixel grayscale images of handwritten single digits between 0 and 9, SVHN is a house numbers digit dataset with $32 \times 32$ images of 10 classes and CIFAR-100  has $32 \times 32$ images with 100 classes. 
We follow a setting where we initially train the model for \emph{half the number of classes}\cite{icarl, lucir, pod} \emph{and then incrementally add 2, 5 and 10 classes in each task} for ImageNet and CIFAR-100 experiments. A strict memory budget is considered where only 20 exemplars per class are stored. 
For MNIST and SVHN experiments, we always add 2 classes per task with a fixed exemplar memory of 4.4k as followed in \cite{rps}.

\subsection{Implementation Details}

\noindent \textbf{Small-scale Datasets: } NesT-tiny architecture with a configuration of $S = 1$ is used for CIFAR-100 experiments, while for SVHN and MNIST we use $S = 2$.
The embedding dimension is set to 192, the number of hierarchy levels is 3, the number of transformer encoder blocks per level is 4 and the number of heads in each level is 6. Augmentations such as random erasing, cutmix, mixup and random augment are used as suggested in \cite{nest}. However, mixup is not used in the incremental phases. The suitable choices of hyper-parameters found empirically are  $\lambda$ = 7, $\tau$ = 1 and $\gamma$ = 0.1. We use a batch size of $128$ and observe that performing distillation only over memory samples is more favorable.

\noindent \textbf{ImageNet:} We use NesT-tiny architecture for ImageNet experiments too. We set $S = 4$, embedding dimensions is set to $(96,192, 384)$, the number of hierarchy levels are 3, the number of transformer encoder blocks per level are $(3, 6, 12)$ and the number of heads per level are $(2, 2, 8)$. Augmentations such as random erasing, cutmix, mixup and random augment are used as suggested in \cite{nest}. Mixup is also used in the incremental phases. Empirically, we find that the hyper-parameters when set to $\lambda$ = 4, $\tau$ = 0.3 and $\gamma$ = 0.05 yield the best results. 
We observe that performing feature distillation over all samples provides more stability when training NesT on ImageNet. We use a batch size of 384 for ImageNet-100 and 1024 for ImageNet-1K.\\
For both small and large scale datasets, the model is trained for 250 epochs, 150 epochs in case of 2 classes per phase. Weighted Adam~\cite{adamw} is used as the optimizer. The learning rate starts from $2.5\text{e}-4$ and decays following cosine annealing scheduler. We make use of PyTorch implementation of NesT from timm library~\cite{rw2019timm} and train on an RTX A6000 GPU.

\subsection{Results}
We conduct exhaustive experimental analysis to test the mettle of our approach. We use three metrics to quantify the performance: 1) average accuracy across all phases, 2) accuracy of the last phase and 3) forgetting rate $\mathcal{F}$ defined as the difference between accuracy of $\bm{\Theta_0}$ and $\bm{\Theta_N}$ on the same test data $\mathcal{D}^{test}_0$ following \cite{mnemonic}.  
Further, following \cite{aanet}, we either use the softmax predictions from the final classifier or use a nearest class mean based classifier \cite{icarl} during inference. We refer to these as \Ours and \Ours-NCM respectively in the results. Note that the bold and underlined values in the tables indicate the best and second best metrics respectively.

\noindent \textbf{CIFAR-100: } Tab. \ref{table_cifar} and Fig.~\ref{plot} summarizes the results on CIFAR-100 dataset when we add incrementally add $10$, $5$ and $2$ classes respectively to a model trained on the first $50$ classes. We observe that as the number of phases increases, the gap between \Ours and the compared methods progressively increases -- thanks to our dual-distillation and logit-correction mechanisms. Specifically, for 25 task experiment, our method improves average accuracy from $62.83\%$ to $68.68\%$ (+{$5.8\%$}).
This was achieved using NesT-tiny architecture, which has 6.2 million parameters compared to DyTox's 10.73 million parameters.

\noindent \textbf{ImageNet: }
We summarize the results of incrementally learning ImageNet Subset-100 dataset in Tab.~\ref{table_imagenet100}. We consider $5$, $10$ and $25$ task incremental setting. Our method achieves the best average accuracy of $77.5\%$ in the $5$ phases settings and $72.43\%$ in $25$ phases settings and is comparable to \cite{pod,dytox} in $10$ phases setting. Tab.~\ref{table_imagenet1k} summarizes ImageNet-1K results in $5$ and $10$ phase setting. Unlike small scale datasets, ImageNet shows relatively better performance while using NCM. The aforementioned behaviour is not present in previous CNN based methods~\cite{pod}. This can be attributed to two factors: first, transformers have better generalization compared to CNNs \cite{delveds}, which results in better class means, second, NesT uses higher embedding dimension for large scale datasets which can help in NCM based classification.

\noindent \textbf{MNIST, SVHN: } Thanks to the better inductive biases of our hybrid architecture, \Ours can scale to small datasets like MNIST and SVHN too. This uniquely differentiates us to recent efforts~\cite{dytox,impr} in utilizing transformer architecture for incremental learning. Tab. \ref{table_small} summarizes the average accuracy results on these datasets by adding two new classes in every incremental phase. Our method clearly surpasses other methods by more than $2\%$ for MNIST and $5\%$ for SVHN dataset.
\begin{table}
\centering

\begin{tabular}{p{4cm}p{1.5cm}p{1.5cm}}

  \toprule
    Method & MNIST & SVHN\\
    \midrule
    \midrule
    EWC \cite{ewc} & 19.80 & 18.21\\
    LwF \cite{lwf} & 24.17 & -\\
    GEM$ ^\ast$ \cite{gem} & 92.20 & 75.61\\
    RPS-Net$ ^\ast$ \cite{rps} & 96.16 & 90.83\\
    \midrule
    \rowcolor{blue!6} \textbf{\Ours$ ^\ast$ (ours)} & \textbf{98.85} & \textbf{95.81}\\
  \bottomrule
\end{tabular}
\caption{Average accuracy (\%) for MNIST, SVHN in 5 tasks setting with 2 classes each with 4.4k fixed memory (\emph{$*$ indicates use of exemplars}}
\label{table_small}
\end{table}

\subsection{Discussions and Analysis}

\begin{figure}[t]
    \centering
    \includegraphics[scale=0.7]{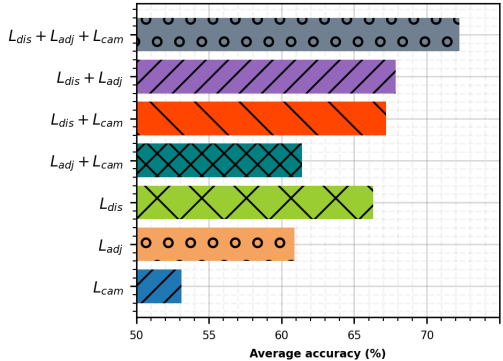}
    \caption{
    We analyze the contribution of each of the constituent component of our loss here. We use a $5$ task CIFAR-100 experiments for this ablation study. 
    }
    \label{fig:loss_abla}
\end{figure}

\customsubsubsection{Contribution from Each Loss Terms: } We analyse the contribution of each component in our loss formulation in  Fig.~\ref{fig:loss_abla}.
We observe that with just cosine distillation, NesT is able to achieve almost comparable accuracy as the baselines \cite{bic,icarl,lucir}. The addition of logit adjustment offset alone brings about $1.5\%$ - $2\%$ improvement over using cosine distillation loss. We observe that Grad-CAM loss alone is not strong enough to boost the accuracy. This is because of the model's inability to handle abrupt changes in model parameters caused due to class imbalance. However, when combined with other losses, we observe considerable improvement. 

\customsubsubsection{Sensitivity Analysis on $\tau$, $\gamma$, $\lambda$:} There is a trade off between forgetting and learning while doing logit adjustment.
As shown in Tab.~\ref{table_tao}, a high value of $\tau$ effectively reduces the forgetting, but puts much emphasis on old classes that hinders new learning. In contrast, a small value of $\tau$ does not have enough impact on retaining old classes. Table \ref{table_gamma} clearly shows the benefit of using $L_{cam}$ and $L_{dis}$ in improving accuracy. For 5 tasks CIFAR-100 setting, $\tau$=1, $\lambda$=7 and $\gamma$=0.1 obtains the best results.

\begin{table}[htb]
\centering

\begin{tabular}{p{1.5cm}p{1.5cm}p{1.5cm}p{1cm}}

  \toprule
  $\tau$  &
  Avg
  $\uparrow$ &
  Last
  $\uparrow$ & $\mathcal{F}$ $\downarrow$\\
    \midrule
    \midrule
    0 & 60.26 & 48.51 & 38.41\\
    0.5 & 66.93 & 57.65 & 28.84\\
    0.75 & 69.34 & 60.29 & 22.67\\
    1 & 72.21 & 66.30  & 12.09\\
    1.25 & 71.72 & 65.61  & 11.32\\
    1.5 & 71.14 & 65.07  & 07.70\\
  \bottomrule
\end{tabular}
\caption{Effect of varying $\tau$ in a $5$ task CIFAR-100 setting.}
\label{table_tao}
\end{table}

\begin{table}[htb]
\centering

\begin{tabular}{p{1cm}p{1cm}p{1cm}p{0.5cm}p{0.5cm}p{1cm}p{1cm}}
  \toprule
  \textbf{$\gamma$}  &
  Avg
  $\uparrow$ &
  Last
  $\uparrow$ & & \textbf{$\lambda$} & Avg  $\uparrow$ & Last $\uparrow$\\
    \midrule
    \midrule
    0 & 67.85 & 60.16 & & 0 & 57.34 & 48.67\\
    0.05 & 71.97 & 66.39  & & 5 & 71.78 & 66.21\\
    0.10 & 72.21 & 66.24 & & 7 & 72.35 & 66.36\\
    0.15 & 72.03 & 66.25 & & 9 & 72.17 & 66.57\\
    0.20 & 71.81 & 66.16 & & 12 & 71.95 & 66.46\\
  \bottomrule
\end{tabular}
\caption{Effect of $\gamma$ and $\lambda$ in a $5$ task CIFAR-100 setting.}
\label{table_gamma}
\end{table}

\customsubsubsection{On Data Used for Distillation: }
We study the effect of distilling from exemplars verses all the data-points here. Applying distillation on all data combined with debiasing techniques such as logit adjustment, could impede learning of new tasks. Although it helps in reducing catastrophic forgetting, it adds a lot of constraints on the learning of new classes. This  becomes more  prominent in case of small datasets, due to less number of learnable parameters. Tab.~\ref{table_mix} shows the positive effect of only applying distillation on exemplars, which is intuitive.

\customsubsubsection{Effect of Mixup:} 
Our method uses mixup augmentation \cite{zhang2017mixup} in the initial phase where half of the classes are learnt. However, we observe differences in performance when using mixup in incremental phases. For CIFAR-100, using mixup in incremental phases proves to be unfavorable. This is because distillation loss is indeed adding strong regularization for these small scale datasets. We see this trend in Tab.~\ref{table_mix}.

\begin{table}[htb]
\centering
\begin{tabular}{p{1cm}p{2.5cm}p{1cm}p{1cm}}
  \toprule
  Mixup  & 
  Distillation & $S=1$ & $S=2$\\
    \midrule
    \midrule
    \checkmark & all samples & 62.10 & 60.80\\
                & all samples & 71.87  & 65.45\\
    \checkmark & exemplars & 66.71 & 64.71\\
                & exemplars & 72.21  & 67.07\\
  \bottomrule
  
\end{tabular}
\caption{
   Effect of using Mixup and distillation in incremental phases.
Impact of different patch size $S$ is also shown. The average accuracy for $5$ task CIFAR-100 is reported.
}
\label{table_mix}

\end{table}

\customsubsubsection{Generality of our Approach: }
We note that our proposed loss formulation ($L_{dis}$, $L_{cam}$ and $L_{adj}$) is agnostic to the backbone network being used. To elucidate this, we swap the NesT backbone with a standard ResNet-18 backbone and report the result in Tab.~\ref{tab:generalizability} for $5$ task Imagenet-100 setting. We borrow the hyper-parameters for the ResNet backbone from AANet \cite{aanet} and use $\tau$=0.3, $\lambda$=5 and $\gamma$=0.01.
This shows that our proposed distillation and logit adjustments helps in reducing forgetting, however forgetting is much higher when compared to \Ours.

\begin{table}[ht]
\centering
\begin{tabular}{p{4cm}p{1cm}p{1.1cm}p{1.1cm}}
  \toprule
  {Setting}  & {Avg $\uparrow$} & {Last $\uparrow$} & {$\mathcal{F} \downarrow$}\\
    \midrule
    \midrule
    ResNet + $L_{dis}$ & 68.52 & 55.83 & 34.81\\
    ResNet + $L_{dis}$+$L_{adj}$ & 71.84 & 61.81 & 24.25\\
    ResNet + $L_{dis}$+$L_{adj}$+$L_{cam}$ & 71.97 & 62.26 & 23.18\\
  \bottomrule
\end{tabular}
\caption{Our proposed loss applied to a ResNet-18 backbone on 5 tasks ImageNet-100 setting, improvement in performance is still observed. }
\label{tab:generalizability}

\end{table}

\section{Conclusion}
We propose \Ours, a hybrid ViT based model that is tuned for class incremental learning. We propose two fundamental components to effectively balance the stability and plasticity required for a continual learner: 
First, we view each incremental phase as a long tail distribution and show the effectiveness of a simple logit offset in reducing inherent bias towards new classes. Second, we show that 
preserving the spatial attention response of a model via distillation can help in improving the spatial awareness of the model and reduce catastrophic forgetting. \Ours achieves superior performance gains over the state-of-the-art methods on MNIST, SVHN, CIFAR-100 and ImageNet. We hope our approach can serve as a simple baseline for incremental hybrid ViTs.

{\small
\bibliographystyle{ieee_fullname}
\bibliography{egbib}
}

\newpage

\begin{center}
  \section*{\Ours: Supplementary Materials}  
\end{center}

\subsection*{Training Details}

\emph{\bf{Zeroth phase}}: The zeroth phase training starts with 10 warm up epochs for small scale datasets, and 20 epochs for large scale datasets. 

\emph{\bf{Incremental phases}}: In each incremental phase, the new classes classifier weights are initialized following weight imprinting introduced in \cite{weight_imprint}, old classes classifier weights are  frozen. The learning rate starts from 2.5e-4 for the feature extractor and 2.5e-3 for the classifier. Both learning rates follow a cosine annealing scheduler that decays the weight till it reaches zero at the final epoch. The number of epochs for each phase is 250 in case of 10 classes per task and 5 classes per task, while for 2 classes per task the number of epochs is kept at 150. 

Knowledge distillation factor $\lambda$ is increased every phase as a factor of number of classes as follows:
\begin{equation}
    \lambda_{t}=\lambda_{t-1} \times \sqrt {\frac{B+C}{C}} ,
\end{equation}
Where $B$ is the number of base classes, and $C$ is the number of new added classes every phase.
The classes exemplars are chosen following the same herding method of \cite{icarl}.

\subsection*{Effect of Augmentations}
NesT uses augmentations such as Mixup, RandomErasing and RandAugment. These augmentations have been shown to be useful in stabilizing training and improve performance of hybrid ViTs \cite{swin, nest}. The importance of these augmentations has also been discussed in the NesT paper. We show the effect of these augmentations when used in the incremental phases.

\begin{table}[h]
\centering
\caption{Effect of augmentations when used in incremental phases of 5 tasks setting for CIFAR100}
\begin{tabular}{p{5cm}p{1.5cm}}
  \toprule
  \textbf{Augmentations}  & \textbf{Average accuracy}\\
    \midrule
    \midrule
    With Mixup & 71.85\\
    With Randaug, RandomErasing & 71.82\\
    With all augmentations & 72.23\\
  \bottomrule
\end{tabular}
\label{table_aug}
\end{table}

\subsection*{Qualitative analysis}

\begin{figure*}[htp]
    \centering
    \includegraphics[width=\textwidth]{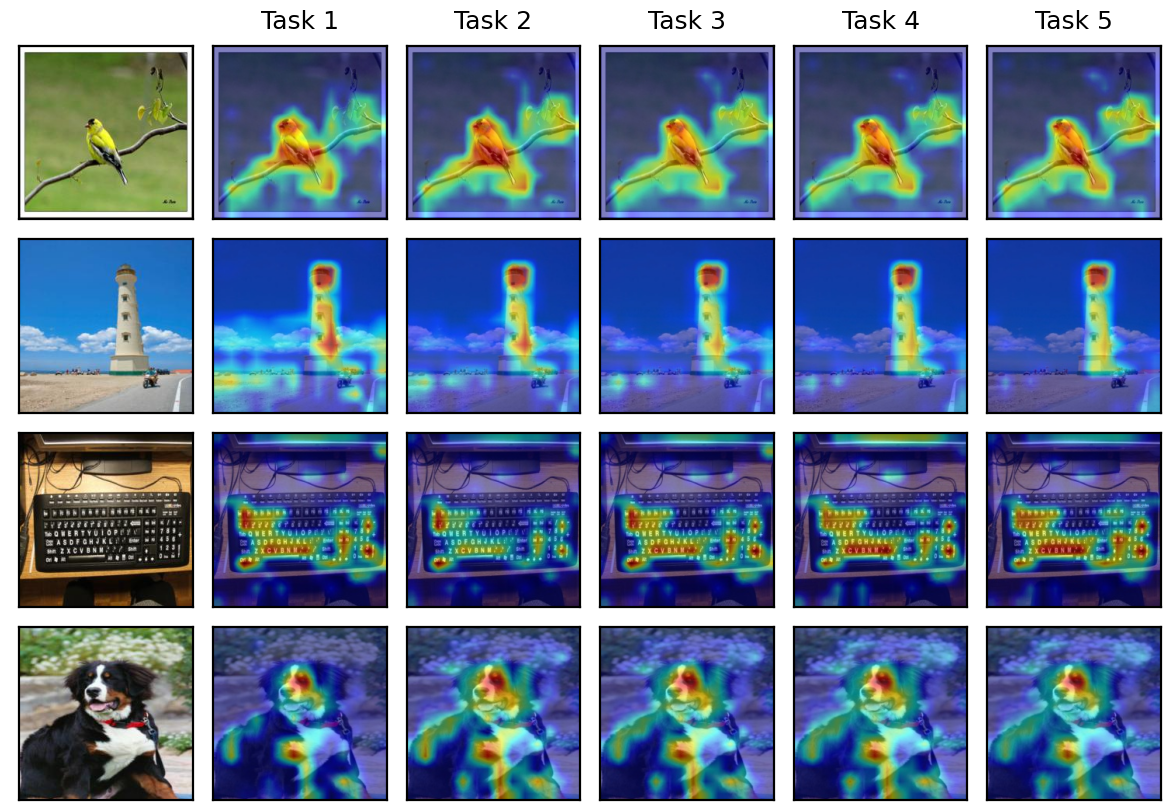}
    \caption{Grad-CAMs for images from ImageNet subset-100 as incremental learning progresses. This shows that Grad-CAM distillation helps $\textrm{D}^3\textrm{Former}$ maintain attention on discriminative patches. (\emph{figure best viewed with zoom-in})}
    \label{fig:gradcam}
\end{figure*}

Figure \ref{fig:gradcam} shows some qualitative results in the form of Grad-CAMs with increasing number of incremental tasks. It is observed that the model does not forget much and makes use of the discriminatory regions in an image to make the correct prediction.

\end{document}